\pdfoutput=1

\documentclass[11pt]{article}

\usepackage[preprint]{acl}

\usepackage{times}
\usepackage{latexsym}

\usepackage[T1]{fontenc}

\usepackage[utf8]{inputenc}

\usepackage{microtype}

\usepackage{inconsolata}

\usepackage{amsmath}
\usepackage{amssymb}
\usepackage{mathtools}
\usepackage{amsthm}

\usepackage{graphicx}
\usepackage{subfigure}

\usepackage[capitalize]{cleveref}

\usepackage{booktabs} %
\usepackage{multirow}
\usepackage{microtype}
\usepackage{hyperref}
\usepackage{xurl}

\usepackage{algorithm}
\usepackage{listings} %

\usepackage{xspace}

\newcommand{\relayattn}{RelayAttention\xspace}

\makeatletter
\DeclareRobustCommand\onedot{\futurelet\@let@token\@onedot}
\def\@onedot{\ifx\@let@token.\else.\null\fi\xspace}

\def\eg{\emph{e.g}\onedot} 
\def\ie{\emph{i.e}\onedot}

\def\wrt{w.r.t\onedot}

\makeatother

\RequirePackage{enumitem}
\setlist[itemize]{noitemsep,leftmargin=*,topsep=0em}
\setlist[enumerate]{noitemsep,leftmargin=*,topsep=0em}

\newcommand{\impv}[1]{(\textcolor{blue}{$\uparrow$#1\%})}
\newcommand{\decs}[1]{(\textcolor{blue}{$\downarrow$#1\%})}
\newcommand{\code}[1]{{\texttt{\small #1}}}

\title{\emph{RelayAttention} for Efficient Large Language Model Serving \\
	  with Long System Prompts}

\author{
Lei Zhu\textsuperscript{1} \quad 
Xinjiang Wang\textsuperscript{2} \quad
Wayne Zhang\textsuperscript{2} \quad 
Rynson Lau$^{1\dagger}$ \quad \\
 \textsuperscript{1} City University of Hong Kong
 \qquad \textsuperscript{2} SenseTime Research \\
 {\tt\small {lzhu68-c@my.cityu.edu.hk}, \{wangxinjiang,wayne.zhang\}@sensetime.com,  Rynson.Lau@cityu.edu.hk} \\
 {\tt\small Code: \url{https://github.com/rayleizhu/vllm-ra}}
}

\begin{document}
\maketitle

\newcommand\blfootnote[1]{%
\begingroup
\renewcommand\thefootnote{}\footnote{#1}%
\addtocounter{footnote}{-1}%
\endgroup
}
\blfootnote{$^\dagger$ Corresponding author.}

\begin{abstract}
A practical large language model (LLM) service may involve a long system prompt, which specifies the instructions, examples, and knowledge documents of the task and is reused across requests.
However, the long system prompt causes throughput/latency bottlenecks as the cost of generating the next token grows \wrt the sequence length.
This paper aims to improve the efficiency of LLM services that involve long system prompts.
Our key observation is that handling these system prompts requires heavily redundant memory accesses in existing causal attention computation algorithms.
Specifically, for batched requests, the cached hidden states (\ie, key-value pairs) of system prompts are transferred from off-chip DRAM to on-chip SRAM multiple times, each corresponding to an individual request.
To eliminate such a redundancy, we propose \relayattn, an attention algorithm that allows reading these hidden states from DRAM exactly once for a batch of input tokens.
\relayattn is a free lunch: it maintains the generation quality while requiring no model retraining, as it is based on a mathematical reformulation of causal attention.
We have observed significant performance improvements to a production-level system, vLLM, through integration with RelayAttention. The improvements are even more profound with longer system prompts.
\end{abstract}

\section{Introduction}
\label{sec:intro}

\begin{figure}
    \centering
    \includegraphics[width=0.9\linewidth]{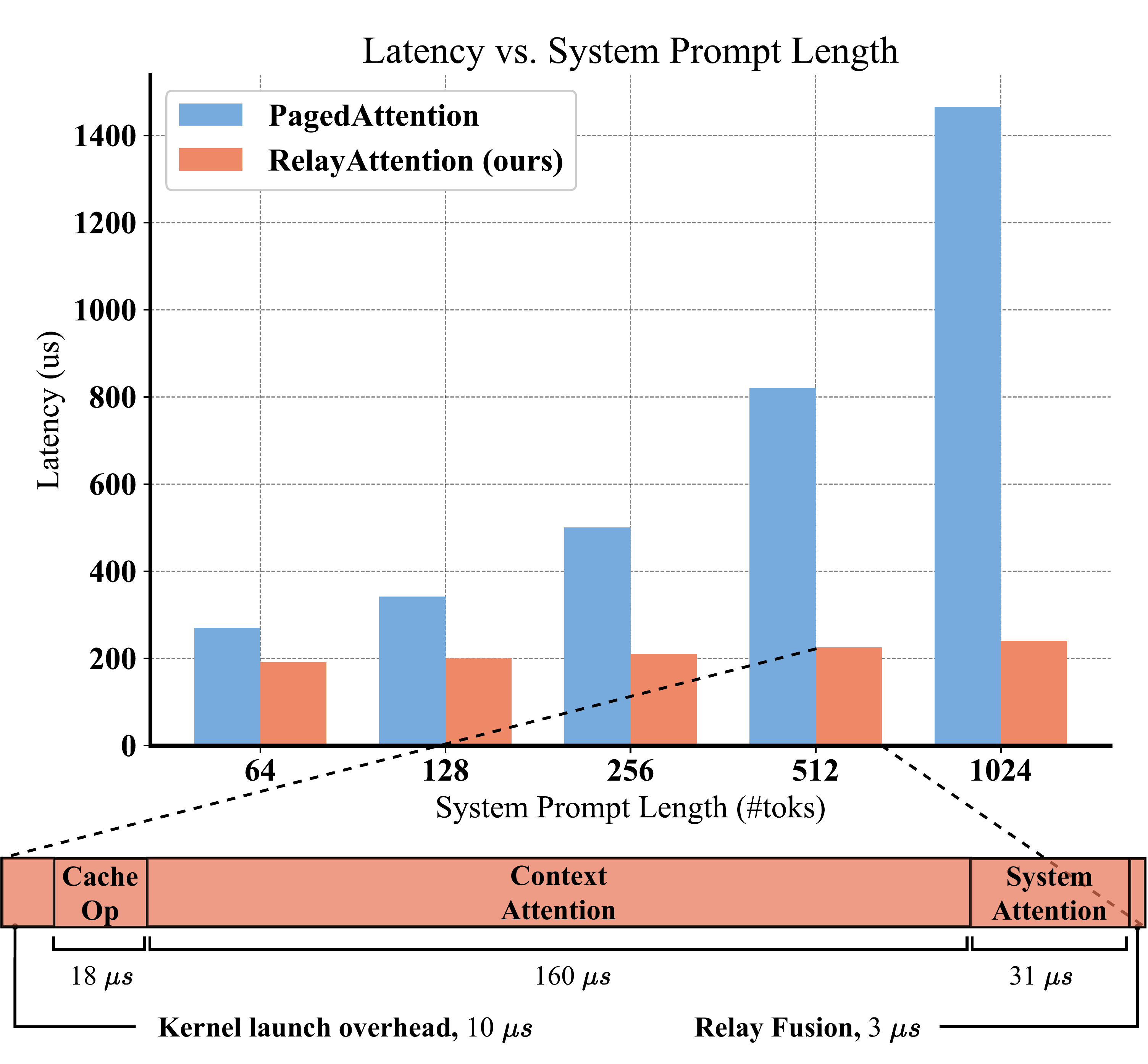}
    \caption{Llama-30B attention inference latency \wrt system prompt length (A40 GPU, batch size 32). We set the length of (request-specific) contexts, which include user prompts and previously generated tokens, to 128. 
    }
    \label{fig:profiling}
    \vspace{-4mm}
\end{figure}

After around one decade of rapid development~\cite{sutskever2014seq2seq,vaswani2017attention,radford2018gpt1,openai2023gpt4}, we have experienced a revolution of large language models (LLMs) over the past year.
LLMs like GPT-4~\cite{openai2023gpt4} and Gemini~\cite{GeminiGo29} are so powerful that they can now serve as programming copilots~\cite{chen2021evaluating,GitHubCopilot}, universal chatbots~\cite{googlebard,openaichatgpt}, computer assistants~\cite{WindowsCopilot} and other roles that penetrate our daily life.
However, the high inference cost of these large models has become a substantial obstacle to serving more people~\cite{kwon2023vllm}.
It is therefore important to improve the hardware utilization so that LLMs can have a higher throughput within a fixed hardware budget.

\begin{figure}
    \centering
    \includegraphics[width=0.98\linewidth]{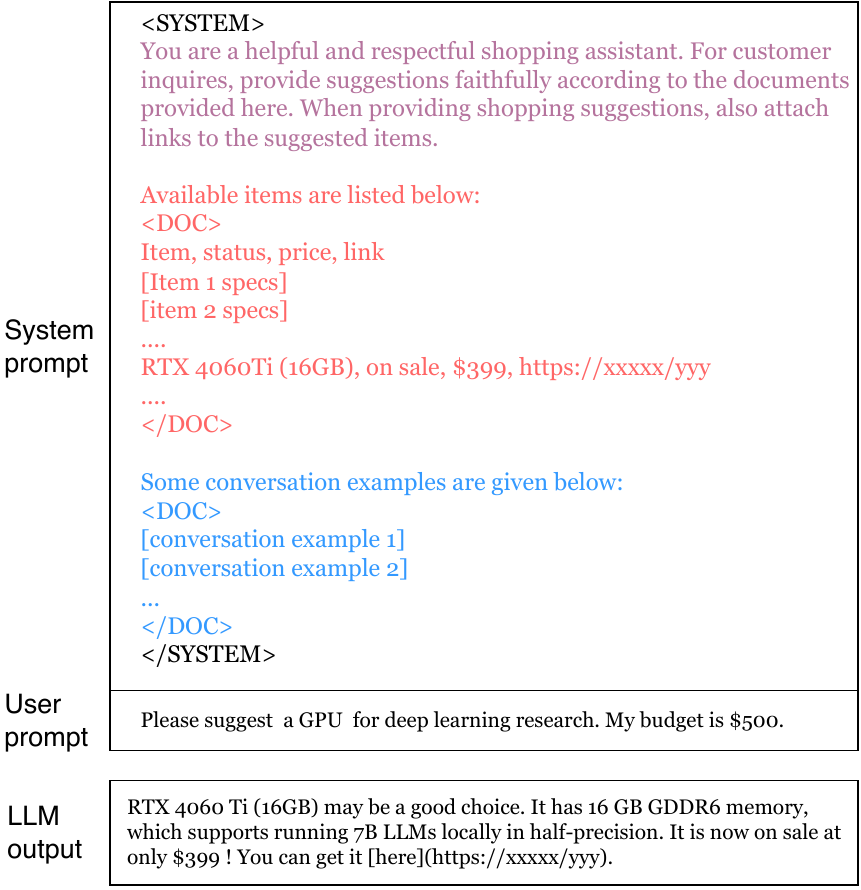}
    \caption{A system prompt may include \textcolor[HTML]{B5739D}{instructions}, \textcolor[HTML]{FF6666}{knowledge documents} and \textcolor[HTML]{3399FF}{few-shot examples}. Here, we use the shopping assistant as an example application.}
    \label{fig:shopping}
    \vspace{-3mm}
\end{figure}

LLM services commonly use an application-specific system prompt~\cite{systemprompt} to specify the task's instructions.
The system prompt is concatenated with the user prompt as the full input to the LLM for response generation and is shared by all requests to a service.
The system prompt becomes long if the service provider wants to provide detailed guidelines and examples for better response quality or apply more restrictions/policies for ethical safety.
As the sequence length that LLMs can process grows~\cite{Claude,chen2023longlora,deepseekai2024deepseek}, some emerging professional applications, such as legal analysis~\cite{cui2023chatlaw,nay2023large}, healthcare applications~\cite{steinberg2021language,rasmy2021med}, and the shopping assistant example shown in \cref{fig:shopping}, may include one or more knowledge documents to provide domain-specific knowledge, resulting in even longer system prompts. 
Although long system prompts are beneficial to improving the generation quality or enabling new applications, they also pose a challenge to the LLM service: the inference throughput and latency of the service can be heavily degraded, thus increasing the per-request cost.
This is inherently caused by the causal attention, in which each new token is generated by ``looking at'' \emph{all precedent} ones.

In this paper, we propose a novel approach to mitigate the efficiency problem of using long system prompts in LLM services.
Our key observation is that there are not only redundant \emph{memory footprint}~\cite{kwon2023vllm} and computations~\cite{gim2023promptcache} corresponding to the system prompt, but also unnecessary \emph{memory accesses} during causal attention computation.
Specifically, while the system prompt is shared by all requests, its hidden states (\ie, key-value pairs) are read from DRAM multiple times by existing attention algorithms such as PagedAttention~\cite{kwon2023vllm} and FlashAttention~\cite{dao2022flashattention,dao2023flashattention2}, each for an individual request in the batch.
This severely slows down LLM inferences, which are known to be memory-bound (\cref{sec:method_bottleneck}).
To eliminate such redundant memory access, we propose \relayattn, an exact algorithm to compute causal attention based on a mathematical reformulation of it.
The key idea of \relayattn is to group the matrix-vector multiplications corresponding to the system prompt into matrix-matrix multiplications, which allow loading the hidden states of the system prompt from DRAM exactly once for all request tokens in a batch (\cref{sec:method_relayattn}).
As a result, the attention inference latency grows much slower than PagedAttention \wrt the length of system prompt, as shown in \cref{fig:profiling}.
We provide an in-depth analysis of the theoretic speedup of the standalone attention based on the IO redundancy reduction (\cref{sec:method_speedup}).
Our empirical results for end-to-end serving further verify the efficiency: integrating \relayattn into vLLM~\cite{kwon2023vllm}, an already highly optimized production-level LLM serving system, we still observe up to $2.2\times$ sustainable request rate and $2.0\times$ throughput with the Llama2-7B model for a chatbot workload. 
Similar efficiency improvements are also observed for several other popular LLMs and are consistent on several data center GPUs.
The efficiency gains continue growing with longer system prompts.

Our key contributions can be summarized as:
\begin{itemize}
    \item We have identified a LLM service bottleneck that has not been studied by existing works: there are highly redundant \emph{memory accesses} caused by long system prompts.
    We anticipate that our analysis will inspire more works on deep architectures with IO-awareness~\cite{dao2022flashattention,gu2023mamba}.
	\item We propose \relayattn, a novel approach to compute exact causal attention. It allows accessing cached hidden states of the system prompt exactly once for a batch of request tokens. We conduct an in-depth analysis of the theoretic speedup brought by \relayattn.
    \item We empirically verify the end-to-end efficiency improvement by integrating \relayattn into vLLM, a production level LLM serving system, and observe non-trivial efficiency gains on several popular LLMs with different GPUs.
\end{itemize}

\section{Related Works}
\label{sec:relworks}

Our approach aims to improve the inference efficiency of transformer-based LLMs (\cref{sec:relworks_llm}).
It is based on extending the widely used Key-Value Cache mechanism (\cref{sec:relworks_kvcache}). 
We also briefly review other techniques for accelerating LLM inference, which may complement ours (\cref{sec:relworks_inference}).

\subsection{Inference of Transformer-based LLMs}
\label{sec:relworks_llm}

The inference of these transformer-based LLMs follows the iterative next-token-prediction paradigm.
Specifically, the next token is generated in each time step by attending to all precedent tokens.
The generated token is then appended to the end of the current sequence. The generation then continues until a stopping criterion (\eg, the new token is \code{<eos>}, which indicates the end of the sequence) is met.
A basic approach to implementing such a generation procedure is to perform full self-attention with a casual mask over the entire up-to-present sequence at each time step, just as we do while training the model~\cite{radford2018gpt1}.
This way, a single generation step takes a quadratic complexity \wrt the length of the up-to-present sequence.
Next, we will look at how this complexity can be reduced to linear using the Key-Value Cache.

\subsection{Key-Value Cache}
\label{sec:relworks_kvcache}

Based on the observation that historical tokens are not affected by the future ones during LLM decoding, Key-Value (KV) Cache avoids repetitive computation of the hidden key-value pairs (KVs) by caching them on the fly and then reusing the cached KVs in every subsequent steps~\cite{yu2022orca,pope2023efficiently}.
With KV Cache, in each time step, only a \emph{single token} (\ie, the latest generated one) is used as the query, and the next token is produced by attending to the cached KVs.
The generation complexity thus reduces from quadratic to linear \wrt the up-to-date sequence length.

Some recent research further accelerates LLM inferences by pruning superfluous KV cache data~\cite{zhang2023h2o} or compressing it~\cite{liu2023cachegen} to reduce key-value pairs to be cached.
However, these approaches introduce algebraic discrepancies between model training and inference. Hence, they may hurt the generation quality and/or require extra finetuning efforts.
In contrast, our approach maintains generation quality and is plug-and-play, as it is based on a mathematical reformulation of causal attention.
The acceleration of our approach comes from reducing redundant memory access of the KV cache.
Therefore, it is orthogonal and complementary to prefix sharing in PagedAttention~\cite{kwon2023vllm}, which eliminates redundant \emph{memory footprint} of system prompts, and
is unlike PromptCache~\cite{gim2023promptcache}, which eliminates the redundant \emph{computation} of the reusable prefix KVs and thus only accelerates the prompt phase (\cref{sec:method_bottleneck}).

\subsection{Other Optimizations for LLM Inference}
\label{sec:relworks_inference}

Besides the KV Cache, several other techniques optimize LLM inference in a post-training manner.
For example, network quantization techniques can also be applied to LLMs as they are architecture-agnostic, even though they may need some adaptations to improve the generation stability and quality~\cite{frantar2022gptq,xiao2023smoothquant,lin2023awq}.
FlashAttention~\cite{dao2022flashattention,dao2023flashattention2} is another technique to optimize LLMs' throughputs on GPUs by avoiding redundant write/read of attention probability matrix into/from DRAM.
A production-level LLM serving system may also include continuous batching~\cite{yu2022orca}, which enables iteration-level scheduling of requests, and speculative sampling~\cite{chen2023speculative_a,leviathan2023speculative_b}, which uses a smaller model to generate a draft and then uses the large model to check and correct it. 
Our approach can work together with these components with no conflicts.

\begin{figure*}[ht]
    \centering
    \includegraphics[width=1\linewidth]{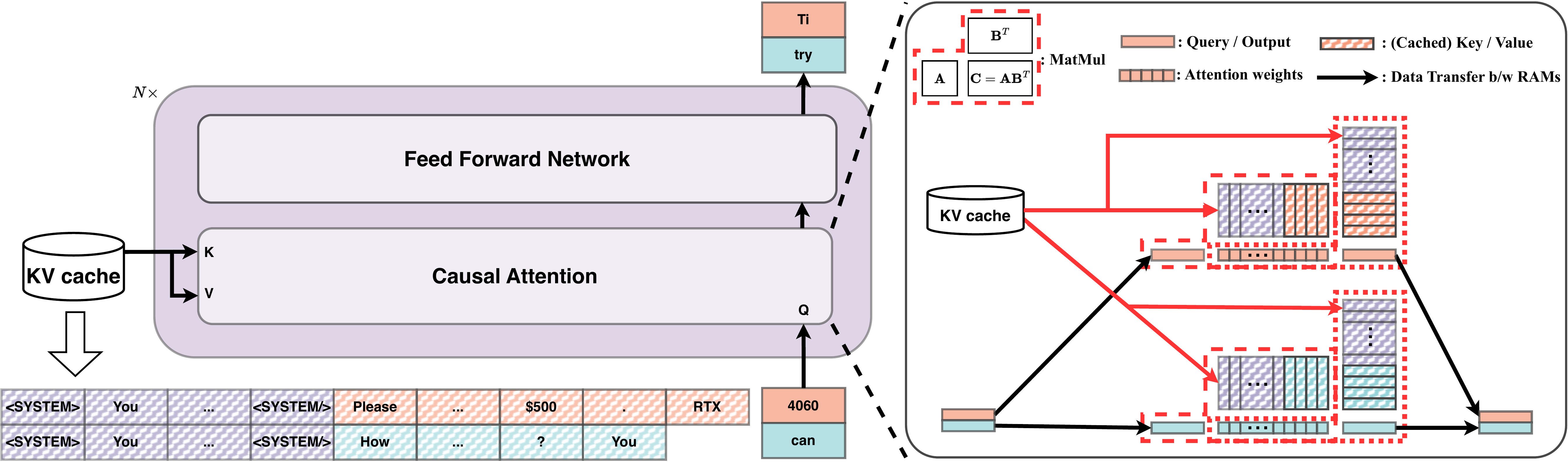}
    \vspace{-6mm}
    \caption{A decoding step during the autoregressive generation phase. On the right side, we provide a closer view of the attention computation with IO-awareness. Note that the floating operations are executed in the fast on-chip SRAM, while the KVs are cached in the slow off-chip DRAM.
    As highlighted with the dashed boxes and red arrows, (1) the computation mainly involves matrix-vector multiplications; and (2) while being shared by all requests, the system KVs are transferred from DRAM to SRAM multiple times, each for one request.
    }
    \label{fig:gemv_decoding}
    \vspace{-2mm}
\end{figure*}

\section{Methodology}
\label{sec:method}

In this section, we elaborate on the proposed approach. 
We begin with a brief preliminary of the hardware utilization of operators in~\cref{sec:method_preliminary}, followed by an analysis of the bottleneck in LLM serving in~\cref{sec:method_bottleneck}, which shows that the redundant memory access slows down the inference especially when the system prompt is long. 
We then introduce \relayattn, a novel algorithm to compute exact causal attention that allows the elimination of the redundancy in~\cref{sec:method_relayattn}.
Finally, we analyze the theoretical acceleration of \relayattn over existing approaches from the perspective of IO-awareness (\cref{sec:method_speedup}).

\subsection{Preliminary: The Latency of Operators}
\label{sec:method_preliminary}

To increase the utilization of arithmetic units, modern processors use pipelining to allow concurrent memory access and computation.
For a perfectly parallelized operator, which maximizes the overlap of data transfer and computation, the runtime latency is determined by the larger one between total memory access time and total computation time.
Given a processor that takes $t_m$ for per-byte access, and $t_c$ for a floating operation on average, the ratio $r$ of the total computation time over the total memory access time for an operator is: %
\begin{equation}
	r = \frac{t_c \times \#\text{floating operations}}{t_m \times \#\text{byte access}} = I \times \frac{t_c}{t_m},
\end{equation}
where $I$ is the arithmetic intensity of the operator:
\begin{equation}
	I = \frac{\#\text{floating operations}}{\#\text{byte access}}.
\end{equation}
When $I < \frac{t_m}{t_c}$, $r$ is less than 1, the operator is memory-bound.
This means that the bottleneck of the operator is memory access, and we can accelerate it only if we can reduce the memory access time.
The speed of modern GPUs far outpaces the bandwidth of its memory (\ie, $\frac{t_c}{t_m} \ll 1$ ), and thus it typically requires a high arithmetic intensity
to achieve full utilization of the computing capability (\eg, A100-SXM4 GPU requires at least 38.2).

For a half-precision (2 bytes/element) general matrix multiplication (GEMM) of problem size $(m, n, k)$: $\mathbf{C} = \mathbf{A}\mathbf{B}^T$,
where $\mathbf{C} \in \mathbb{R}^{m \times n}$,
$\mathbf{A} \in \mathbb{R}^{m \times k}$, 
$\mathbf{B} \in \mathbb{R}^{n \times k}$, the arithmetic intensity is:
\begin{equation}
	I_{gemm} = \frac{2mnk}{2(mk + nk + mn)} < \text{min}\{m, n, k\}.
\end{equation}
When $m, n, k$ are all large (\eg, $>128$), the operation can saturate the utilization of the computing capability due to high arithmetic intensity.
This is normally true for linear projection operations in LLM inference, where $m$ is the number of tokens in a batch, $k$ is the input hidden dimension, and $n$ is the output hidden dimension.
However, as a special case of GEMMs, 
the general matrix-vector product (GEMV) operation, in which there is a vector in $\mathbf{A}$ and $\mathbf{B}$, is always memory-bound as $I_{gemv} < 1$.
This is the case for casual attention computation with cached KVs, as we will show in \cref{sec:method_bottleneck}.

\subsection{Bottleneck of LLM Services}
\label{sec:method_bottleneck}

\begin{figure*}[tp]
    \centering
    \includegraphics[width=1.0\linewidth]{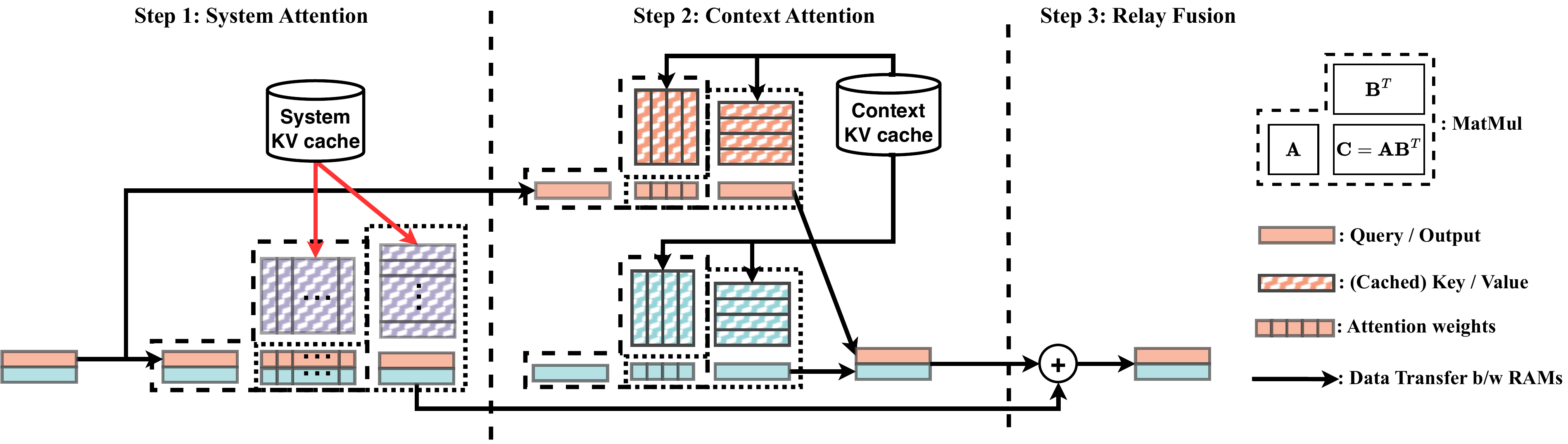}
   \vspace{-5mm}
    \caption{The computation of \relayattn. It is a mathematical reformulation of casual attention in \cref{fig:gemv_decoding},
     but load the System KVs exactly once for a batch of requests (highlighted with red arrows).
    }
    \label{fig:relay_pipeline}
\end{figure*}

Given a batch of user prompts, the LLM inference is usually divided into two phases: 
the \emph{prompt phase}, which computes the hidden states of the full prompts (\ie, the concatenation of system prompt and user prompts) and generates the first new tokens;
and the \emph{autoregressive generation phase}, which generates \emph{all subsequent tokens} sequentially, one token for each request at a time step.
In this work, we focus our investigation on the autoregressive generation phase as it contains the hot spot of response generation\footnote{Besides, the prompt phase can effectively saturate GPU utilization as it
involves large matrix multiplications.}.

In \cref{fig:gemv_decoding}, we demonstrate a time step during the autoregressive generation phase, with the batch size assumed to be 2.
There are two key observations:
\begin{enumerate}
\item \textbf{The computation of attention is memory-bound.}
This is because the attention computation for a request mainly involves two GEMVs (red dashed boxes in \cref{fig:gemv_decoding}), with an arithmetic intensity lower than 1. It thus requires memory access reduction for acceleration.

\item \textbf{There are redundant memory accesses in the typical scenarios where a shared system prompt is prepended to request-specific contexts.}
Specifically, the cached key-value pairs of the shared system prompt (system KVs) are read from off-chip DRAM multiple times, each for a request in the batch (red arrows in \cref{fig:gemv_decoding}).
Such redundancy becomes a substantial overhead when the system prompt is long.
\end{enumerate}

Section~\ref{sec:method_relayattn} proposes the core design of 
\relayattn for removing the redundant memory access.

\subsection{LLM Serving with \relayattn}
\label{sec:method_relayattn}

The key idea of \relayattn is to group multiple matrix-vector multiplications between the batched queries and the cached KVs into single matrix-matrix multiplications, as shown in \cref{fig:relay_pipeline}, allowing system KVs to be read from DRAM exactly once per batch.
\cref{alg:relay_attn} summarizes the algorithm in Pytorch-like pseudo code.
It divides the computation of a causal attention layer into three steps: system attention step, context attention step, and relay fusion step.
In the system attention and context attention steps, we compute two intermediate attention outputs as if the LLM is prompted by the shared system prompt / request-specific context only.
In the relay fusion step, we compute the final output as a convex combination of the two intermediate outputs. Next, we show that \relayattn is computing a mathematical reformulation of casual attention.

\begin{algorithm}[tb]
   \caption{Pseudocode for \relayattn.}
   \label{alg:relay_attn}
   
    \definecolor{codeblue}{rgb}{0.25,0.5,0.5}
    \lstset{
      basicstyle=\fontsize{7.2pt}{7.2pt}\ttfamily\bfseries,
      commentstyle=\fontsize{7.2pt}{7.2pt}\color{codeblue},
      keywordstyle=\fontsize{7.2pt}{7.2pt},
    }
\begin{lstlisting}[language=python]
# INPUT: 
#   q: query tensor for new inputs, (b, m, h, d)
#   k: key tensor for new inputs, (b, m, h, d)
#   v: value tensor for new inputs, (b, m, h, d)
#   kv_cache: context KVs, (N, 2, b, l-s, h, d)
#   sys_kv_cache: sys. KVs, (N, 2, 1, s, h, d)
#   layer_id: the index of current layer, int
#   l_cache: the length of cached key-value, int
# OUTPUT:
#   o: the output of causal attention

# note: (1) we modified the interface of multi-head
# attention to return the log-sum-exp (lse);
# (2) the order of context attention and system 
# attention doesn't matter because of no dependency

# context attention, as if there is no system prompt
# k.size(1) = 1 in autoregressive generation phase
# k.size(1) > 1 in prompt phase
l_new = l_cache + k.size(1)
kv_cache[layer_id, 0, l_cache:l_new, ...] = k
kv_cache[layer_id, 1, l_cache:l_new, ...] = v
o, lse = multihead_attention(
    q, k_cache[layer_id, 0, :l_new, ...],
    v_cache[layer_id, 1, :l_new, ...],
    casual_mask=True)
# system attention
bsz, len, nhead, dim = q.size()
q1 = q.view(1, bsz*len, nhead, dim)
k_sys, v_sys = sys_kv_cache[layer_id].unbind(dim=0)
o_sys, lse_sys = multihead_attention(
    q1, k_sys, v_sys)
o_sys = o_sys.view(bsz, len, nhead, dim)
lse_sys = lse_sys.view(bsz, len, nhead, 1)
# relay fusion
alpha_sys = 1 / (1 + exp(lse - lse_sys))
alpha_usr = 1 - alpha_sys
o = o * alpha_usr + o_sys * alpha_sys 
\end{lstlisting}
\end{algorithm}

Without loss of generality, we consider a single sequence in the batch and a single attention head.
Formally, given an on-the-fly sequence $R$ at generation step $t$, we divide it into three segments
(in order): (1) the system prompt of length $s$, (2) the user prompt of length $u$, and (3) the response generated by the LLM of length $t-1$.
Let $\mathbf{k}_i, \mathbf{v}_i \in \mathbb{R}^{d}$ denote the hidden key, value embedding of the token at position $i \leq l=s+u+t$, and $\mathbf{q}_t \in \mathbb{R}^{d}$ denotes the hidden query embedding in the current step.
The casual attention output $\mathbf{o}_t$ is defined as:
\begin{equation}
\label{eq:casual_attn}
\begin{split}
    \mathbf{o}_t &= \text{Attention}(\mathbf{q}_t, \{\mathbf{k}_i\}_{i=1}^{l}, \{\mathbf{v}_i\}_{i=1}^{l}) \\
                 &= \sum_{j=1}^{l} \frac{\text{exp}(\mathbf{q}_t \mathbf{k}_j^T)}{\sigma_t^{1 \rightarrow l}} \mathbf{v}_j,
\end{split}
\end{equation}
where $\sigma_t^{b \rightarrow e} = \sum_{j=b}^{e} \text{exp}(\mathbf{q}_t \mathbf{k}_j^T)$ is the sum-exp between the start position $b$ and end position $e>b$, associated with $\mathbf{q}_t$. 
By splitting the summation in \cref{eq:casual_attn} at position $s$, which is the end system prompt, we have:
\begin{equation}
\label{eq:split_sum}
\begin{split}
    \mathbf{o}_t &= \sum_{j=1}^{s} \frac{\text{exp}(\mathbf{q}_t \mathbf{k}_j^T)}{\sigma_t^{1 \rightarrow l}} \mathbf{v}_j + 
                    \sum_{j=s+1}^{l} \frac{\text{exp}(\mathbf{q}_t \mathbf{k}_j^T)}{\sigma_t^{1 \rightarrow l}} \mathbf{v}_j.
\end{split}
\end{equation}
Consider the first term on the right side of \cref{eq:split_sum}. As it is close to the $\text{Attention}(\cdot, \cdot, \cdot)$ operation in \cref{eq:casual_attn}, with only a difference in the numerator, it can be rewritten as a rescaled attention:
\begin{equation}
\label{eq:rescale}
\begin{split}
\frac{\sigma_t^{1 \rightarrow s}}{\sigma_t^{1 \rightarrow l}}
\sum_{j=1}^{s} \frac{\text{exp}(\mathbf{q}_t \mathbf{k}_j^T)}{\sigma_t^{1 \rightarrow s}} \mathbf{v}_j.
\end{split}
\end{equation}
This rescaling trick~\cite{milakov2018online,rabe2021memeffattn} can also be applied to the second term on the right side of \cref{eq:split_sum}, and thus $\mathbf{o}_t$ is a convex combination of two scaled attention terms:
\begin{equation}
\label{eq:convex_combination}
\begin{split}
    \mathbf{o}_t =& \alpha_t^{sys}\text{Attention}(\mathbf{q}_t, \{\mathbf{k}_i\}_{i=1}^{s}, \{\mathbf{v}_i\}_{i=1}^{s}) + \\
                  & \alpha_t^{ctx}\text{Attention}(\mathbf{q}_t, \{\mathbf{k}_i\}_{i=s+1}^{l}, \{\mathbf{v}_i\}_{i=s+1}^{l}),
\end{split}
\end{equation}
where $\alpha_t^{sys} = \frac{\sigma_t^{1 \rightarrow s}}{\sigma_t^{1 \rightarrow l}}$ \ and \  $\alpha_t^{ctx}=\frac{\sigma_t^{s+1 \rightarrow l}}{\sigma_t^{1 \rightarrow l}}=1 - \alpha_t^{sys}$ are the combination coefficients.

\noindent
\textbf{Discussion.} Back to the view of a batch, the first term in \cref{eq:convex_combination} for all concurrent requests, namely \emph{system attention}, can be grouped to use large matrix multiplications. This essentially eliminates the redundant access of system KVs as shown in \cref{fig:relay_pipeline}.
In practice, as the sum of exponentials $\sigma_t^{b \rightarrow e}$ is not numerically stable to compute directly, we use the log-sum-exp trick to return $\text{log}(\sigma_t^{b \rightarrow e})$ in attention computation, and the computation of $\alpha_t^{sys}$ is reformulated accordingly in \cref{alg:relay_attn}.
While reformulating the casual attention, we did not assume step $t \neq 1$. This means that \relayattn is also applicable to the prompt phase, where the input of a request is not a single token generated in the last step but contains multiple tokens from the user prompt, as reflected in \cref{alg:relay_attn}.

\begin{figure}
    \centering
    \includegraphics[width=1\linewidth]{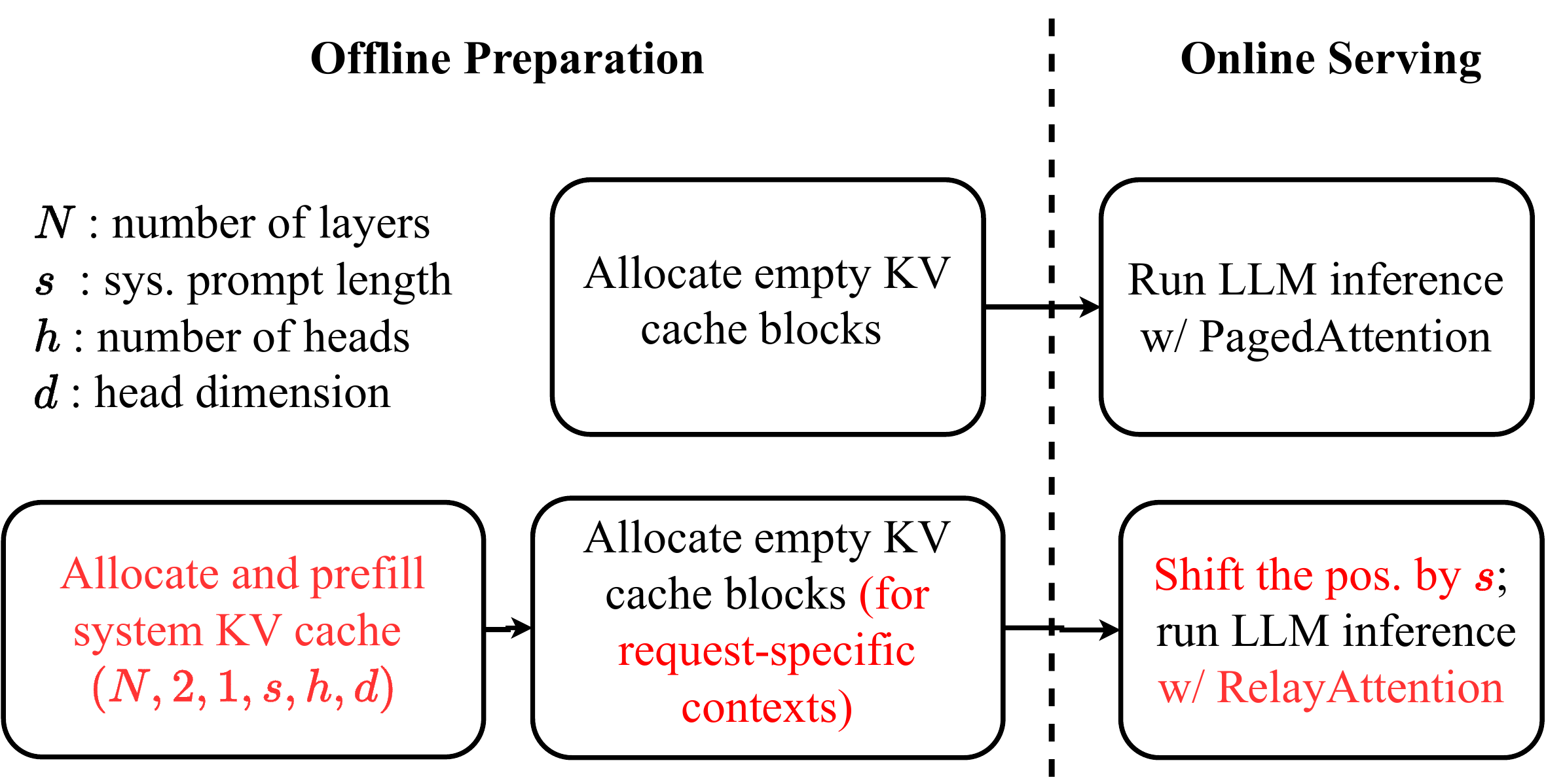}
    \caption{Key modifications (high-lighted in \textcolor[HTML]{FF0000}{red} in the bottom) to integrate \relayattn into an existing LLM serving system (top).}
    \label{fig:system-pipeline}
\end{figure}

\noindent
\textbf{Peripheral adaptations.} There are two major adaptations needed to make \relayattn work better within existing inference systems.
First, instead of using a single KV cache for both the system prompt and the request-specific context, we use a separate \emph{system KV cache} to store system KVs and fill it offline before model serving.
This can be viewed as a combination of prefix sharing in PagedAttention, which eliminates redundant memory footprint of system KVs, and PromptCache~\cite{gim2023promptcache}, which eliminates redundant computation in the \emph{prompt phase}.
Second, as the system KVs are already computed offline, we add an offset of $s$ (\ie, the length of the system prompt) in the position of those request-specific context tokens to make sure of correct position embedding.
\cref{fig:system-pipeline} summarizes the key modifications to integrate \relayattn into an existing LLM serving system (\eg, vLLM~\cite{kwon2023vllm} or TensorRT-LLM~\cite{nvidia2023trtllm}).

\subsection{Theoretical Speedup}
\label{sec:method_speedup}

\begin{figure}
    \centering
    \includegraphics[width=0.98\linewidth]{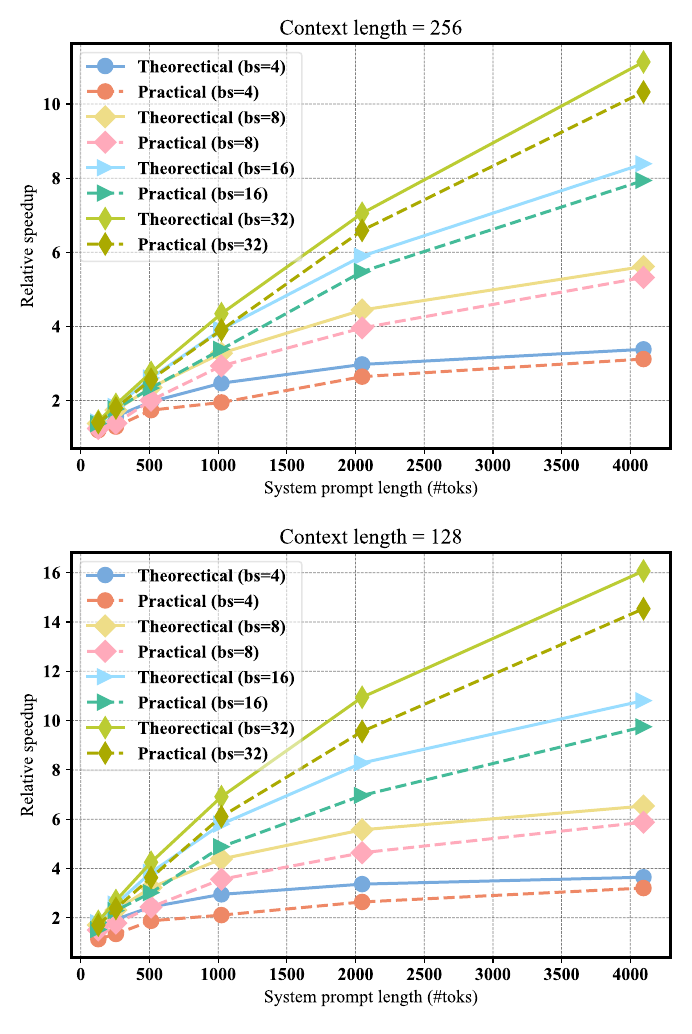}
    \vspace{-3mm}
	\caption{The theoretical and practical speedups for a standalone Llama-30B casual attention with \relayattn. We plot the speedup \wrt the length of the system prompt under two context lengths (128 and 256) and four batch sizes (4, 8, 16, and 32), on an A40 GPU.}
    \vspace{-1mm}
    \label{fig:relayattn_speedup}
\end{figure}

In this section, to derive the theoretical speedup of \relayattn by the memory access reduction, we analyze the memory access during the attention computation of an autoregressive generation step.

Without \relayattn, given a batch of $b$ request tokens, the number of elements $n$ to transfer between DRAM and SRAM is:
\begin{equation}
	n = \underbrace{bd}_{\text{queries}} + \underbrace{b(s+c)d}_{\text{cached KVs}} + \underbrace{bd}_{\text{outputs}},
\end{equation}
where $d$ is the embedding dimension, $s$ is the length of the shared system prompt, and $c$ is the length of request-specific context.
With \relayattn enabled, the number of elements to access $n^\prime$ is:
\begin{equation}
 	n^\prime = \underbrace{(bd + sd + bd)}_{\text{system attention}} + \underbrace{(bd + bcd + bd)}_{\text{context attention}} + \underbrace{3bd}_{\text{relay fusion}}.
\end{equation}
Therefore, the speedup $p$ is:
\begin{equation}
\label{eq:speedup}
	p = \frac{n}{n^\prime} = \frac{s+c+2}{s/b + c + 7}.
\end{equation}
In \cref{fig:relayattn_speedup}, we plot the speedup brought by using \relayattn. The small gaps between the practical and corresponding theoretical curves verify our analysis.

Though the speedup of standalone \relayattn can be analyzed, it is still a question of how an end-to-end LLM serving system can benefit from \relayattn. In \cref{sec:experiment}, we provide an empirical study to answer this question.

\section{Experiments}
\label{sec:experiment}

\begin{figure*}[t]
    \centering
    \includegraphics[width=0.98\linewidth]{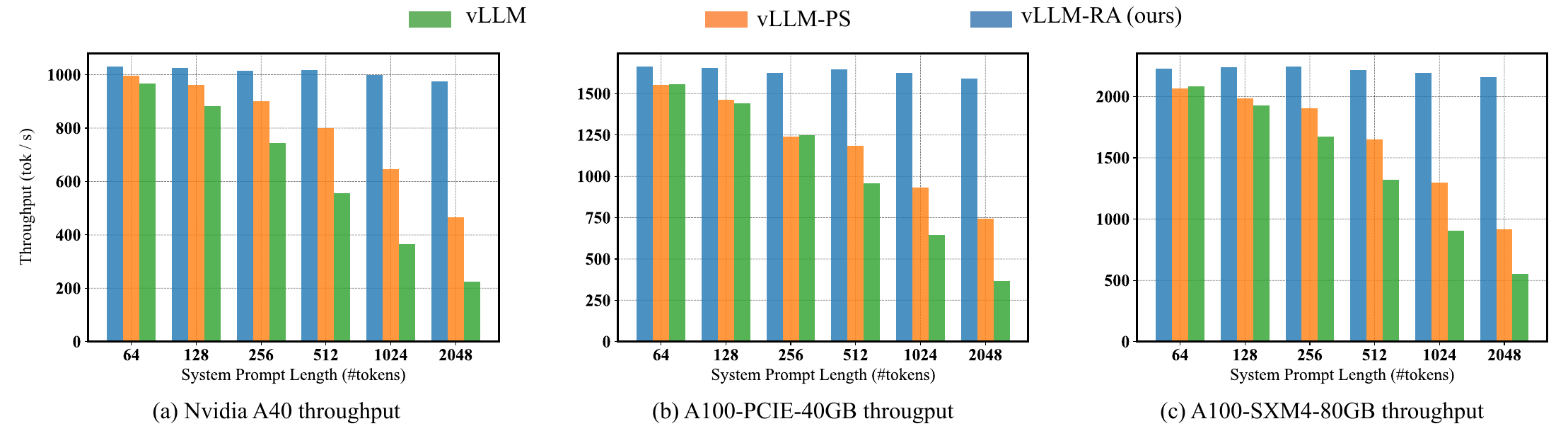}
    \vspace{-1mm}
    \caption{Throughput \wrt system prompt length during the noninteractive processing of ShareGPTv3 dataset.}
    \label{fig:sharegpt-batch}
    \vspace{-1mm}
\end{figure*}

In this section, we conduct experiments to answer the question of how much our approach can help an end-to-end LLM serving system.
We provide the experimental setup in \cref{sec:exp_setup}.
Our major evaluation is conducted with consideration of two scenarios: noninteractive batch processing (\cref{sec:exp_noninteractive}) and interactive service (\cref{sec:exp_interactive}).
We use the Llama2-7B model~\cite{touvron2023llama2} for evaluation unless stated otherwise.
We demonstrate the improvement for more models in \cref{sec:exp_more_models}.

\subsection{Experimental Setup}
\label{sec:exp_setup}

\textbf{Data.} Two datasets are used in our evaluation: ShareGPTv3~\cite{sharegpt} and MMLU~\cite{hendryckstest2021mmlu}.
SharedGPTv3~\cite{sharegpt} contains 53k real conversions between users and ChatGPT~\cite{openaichatgpt}. 
MMLU is a benchmark for measuring massive multitask language understanding in few-shot settings. It consists of 57 tasks covering various subjects and domains, such as mathematics, history, law, and medicine.
Each subject/task contains a series of single-choice questions, and 5 extra questions with answers (as few-shot examples).
The statistics of the benchmarking data are summarized in \cref{tab:data}.

\begin{table}
    \centering
    \setlength{\tabcolsep}{4pt}
    \resizebox{1\linewidth}{!}{
    \begin{tabular}{c|c|c|c}
     \toprule
                 & Sys. prompt len.   & User prompt len. &  Generation len. \\
     \midrule
      ShareGPTv3 & 64-2048       & 4-1024        &  4-2013   \\
      MMLU       & 379-2895      & 55-1219        &  32    \\
      \bottomrule
    \end{tabular}
    }
    \vspace{-2mm}
    \caption{Data for benchmarking. Lengths are measured in token.}
    \label{tab:data}
\end{table}

\vspace{2mm}
\noindent
\textbf{Hardware}. Our experiments involve three GPUs: A40, A100-PCIE-40GB, and A100-SXM4-80GB. However, A40 is used unless stated otherwise. 
The hardware specifications are listed in \cref{tab:gpu}.

\begin{table}
    \centering
    \setlength{\tabcolsep}{4pt}
    \resizebox{1\linewidth}{!}{
    \begin{tabular}{c|c|c|c|c}
     \toprule
                      &  Memory   & Mem. Band.  &  FP16 Peak F.  & Price \\
     \midrule
      A40             & 48 GB   & 696  GB/s     &  37.4 TFLOPs & \$0.40/hr \\
      A100-PCIE-40GB  & 40 GB   & 1,555  GB/s   &  77.9 TFLOPs & \$0.90/hr \\
      A100-SXM4-80GB  & 80 GB   & 2,039  GB/s   &  77.9 TFLOPs & \$1.84/hr \\
      \bottomrule
    \end{tabular}
    }
    \vspace{-2mm}
    \caption{The specifications of the GPUs used in our experiments. Prices are from \href{https://vast.ai/}{vast.ai}.}
    \label{tab:gpu}
    \vspace{-2mm}
\end{table}

\vspace{2mm}
\noindent
\textbf{Three Approaches used for comparison}:%
\begin{itemize}
	\item \textbf{vLLM}~\footnote{\url{https://github.com/vllm-project/vllm}}: a state-of-the-art open-source LLM inference system designed for high throughput LLM serving. We use the v0.2.6 release.
    Note that the core component of vLLM, PagedAttention~\cite{kwon2023vllm}, allows \emph{storing} the shared system KVs exactly once by the prefix sharing technique mentioned in their paper, but this technique is not included in the vLLM v0.2.6 release.
    Considering the importance to save memory for a higher concurrency, we implement a stronger baseline, vLLM-PS as specified below.
    \item \textbf{vLLM-PS}: the augmented version of vLLM implemented by us. It integrates not only prefix sharing but also PromptCache~\cite{gim2023promptcache}, which precomputes the system KVs and reuses them to mitigate the burden of the \emph{prompt phase}.
    Therefore, vLLM-PS eliminates both redundant \emph{memory footprint} and unnecessary \emph{computations} of system KVs.
    \item \textbf{vLLM-RA (ours)}: the modfied vLLM with our \relayattn integrated. Compared with vLLM-PS, this version further eliminates the redundant \emph{memory accesses} of system KVs, as discussed in \cref{sec:method_relayattn}.
\end{itemize}
Note that, vLLM and the two variants mentioned above use dynamic batch size: requests are dynamically batched with varied batch sizes in \{1, 2, 4, 8, 16, 24, 32, 40, ..., 256\} by a scheduler, depending on the available memory and the number of pending requests in the queue. vLLM would schedule as many requests as possible each time to maximize the hardware utilization.

\subsection{Noninteractive Batch Processing}
\label{sec:exp_noninteractive}

For the non-interactive batch processing scenarios
where users just submit their jobs to the LLM services and harvest the processing results later,
we consider the throughput (number of tokens per second) and processing time as the key metrics.

We plot the throughputs \wrt the length of system prompt for processing ShareGPTv3 on the three GPUs in \cref{fig:sharegpt-batch}.
For vLLM, the throughputs degrade heavily as the system prompt becomes long for two reasons:
(1) the system prompt occupies too much memory, and thus heavily limits the batch size/concurrency of decoding;
(2) it takes too much time to handle long system prompts during causal attention computation.
With prefixing sharing, vLLM-PS solves the first problem and achieves up to 108\% improvement on the throughput. 
Our vLLM-RL further solves the second problem and increases the throughput from $1.06\times$ to $4.36\times$ of vLLM.
\cref{tab:mmlu} shows results of the few-shot test on MMLU. We can see that using a long system prompt to include more examples is crucial for improving accuracy.
In the case of the 5-shot test, our vLLM-RA provides a 76\% reduction of processing time on both A40 and A100-SXM4-80GB GPUs.

\begin{table}[t]
    \centering
    \resizebox{1\linewidth}{!}{
    \begin{tabular}{l|c|c|c|c|c}
    \toprule
             & Accuracy   & GPU & vLLM & vLLM-PS & vLLM-RA \\
    \midrule
    \multirow{2}*{1-shot} & \multirow{2}*{37.6\%} & A100-80G & 502  & 336\decs{33} & 306\decs{39} \\
    					  &						  & A40-48G  & 1012 & 675\decs{33} & 621\decs{39} \\
    \midrule
    \multirow{2}*{3-shot} & \multirow{2}*{41.3\%} & A100-80G & 851  & 378\decs{55} & 311\decs{63} \\
                          &					      & A40-48G  & 1751 & 752\decs{57} & 629\decs{64} \\
    \midrule
    \multirow{2}*{5-shot} & \multirow{2}*{43.2\%} & A100-80G & 1308 & 432\decs{67} & 316\decs{76} \\
                          &					      & A40-48G  & 2660 & 850\decs{68} & 641\decs{76} \\
    \bottomrule
    \end{tabular}
    }
    \vspace{-1mm}
    \caption{MMLU few-shot acc. and processing time (s).}
    \label{tab:mmlu}
    \vspace{-1mm}
\end{table}

\subsection{Interactive Serving}
\label{sec:exp_interactive}

\begin{figure*}[t]
    \centering
    \includegraphics[width=0.98\linewidth]{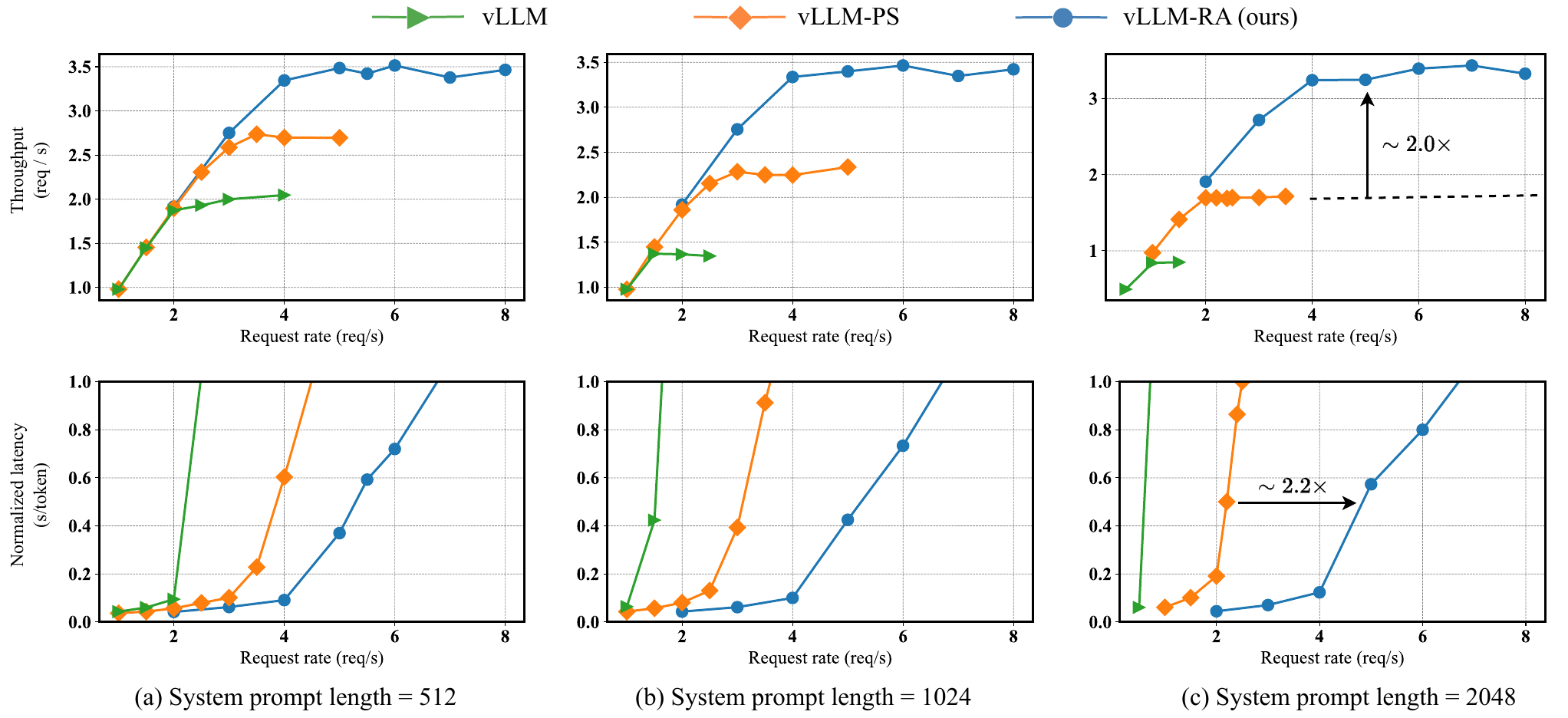}
    \caption{Benchmark interactive serving with requests sampled from the ShareGPTv3 dataset.}
    \label{fig:sharegpt-streaming}
\end{figure*}

An important LLM application is chatbots~\cite{openaichatgpt,googlebard}, in which
interactive LLM services are typically provided.
Unlike the noninteractive scenario, though we still expect a high throughput for good hardware utilization, we also care about the normalized latency (\ie, average per-token latency), which is crucial for user experience.
Following PagedAttention~\cite{kwon2023vllm}, we sample 1000 requests from the ShareGPTv3 dataset to benchmark the efficiency.
The request arrival times are generated using Poisson distribution with different request rates.

As shown in \cref{fig:sharegpt-streaming}, as the request rate increases, the throughput grows gradually until reaching a maximum.
In contrast, the latency remains low at the beginning and then goes up steeply when the highest throughput is achieved.
Around the latency of 0.5s/token, where the user experience and hardware utilization is balanced, vLLM-RA sustains higher request rates than both vLLM and vLLM-PS with clear margins (up to $\sim2.2\times$ when the system prompt length is 2048).

\subsection{The Improvement for More Models}

To verify the efficiency improvement for more models, we choose several other popular LLMs such as Llama2-13B, Llama-30B, Phi-2~\cite{phi2}, and Mistral-7B~\cite{jiang2023mistral} to run the noninteractive batch processing of ShareGPTv3.
As shown in \cref{tab:more-models}, vLLM-RA also provides consistent improvements for these LLMs.

\label{sec:exp_more_models}

\begin{table}
    \centering
    \resizebox{0.98\linewidth}{!}{
    \begin{tabular}{l|c|c|c}
    \toprule
                & vLLM & vLLM-PS & vLLM-RA \\
    \midrule
    \multicolumn{4}{l}{system prompt length = 512} \\
    \midrule
    Llama2-13B  &  0.99 & 1.44 \impv{45} & 1.71 \impv{73} \\
    Llama-30B\textsuperscript{\dag}   & 2.15   & 3.01\impv{40}   & 3.65\impv{70} \\
    Phi-2 (2.7B) & 5.03  & 6.29 \impv{25}  & 8.85\impv{76}  \\
    Mistral-7B   & 3.68  & 5.40 \impv{47}  & 5.90\impv{60}  \\
    \midrule
    \multicolumn{4}{l}{system prompt length = 1024} \\
    \midrule
    Llama2-13B  &  0.66  & 1.23\impv{86} & 1.69\impv{156}  \\
    Llama-30B\textsuperscript{\dag}   & 1.52  & 2.55\impv{68}  & 3.64\impv{139} \\
    Phi-2 (2.7B) & 3.54  & 4.82\impv{36}  & 8.76\impv{147}  \\
    Mistral-7B   & 2.60  & 4.92\impv{89}  & 5.85\impv{125}  \\
    \bottomrule
    \end{tabular}
    }
    \caption{Throughput (req/s) of different models during the batch processing of the ShareGPTv3 dataset. \textsuperscript{\dag}: the 30B model is hosted on two A100-SXM4-80GB GPUs.}
    \label{tab:more-models}
    \vspace{-2mm}
\end{table}

\section{Conclusion}

In this paper, we have identified a bottleneck of using long system prompts in LLM services: there are highly redundant memory accesses corresponding to those system KVs.
We have proposed \relayattn to compute exact causal attention while removing the redundant memory accesses.
An analysis of the theoretical speedup of \relayattn is provided.
Extensive experiments over different GPUs, models, and datasets empirically verify the efficiency gains brought by \relayattn.

\section*{Limitations}

The limitations of \relayattn can be reflected by the theoretical speedup (\cref{eq:speedup}). First, it helps \emph{batched} inference ($b>1$).
The larger the batch size, the more efficient \relayattn is.
When there is only one request, which is the typical case on device-side applications, \relayattn does not help.
Therefore, \relayattn is suitable for cloud-serving scenarios.
Second, when the request-specific contexts (including user prompts and responses) are long (\eg, $2\times$ longer than the shared system prompt), the computation time is dominated by the processing of them; thus the efficiency gain will diminish.
However, as the context length has a long-tailed distribution in many applications (\eg, chatbots), where the majority of user prompts and responses are short, the efficiency gain brought by \relayattn is still considerable.

\section*{Broader Impact Statement}
It is revealed by \relayattn that the inference cost caused by a shared prefix can be amortized by a batch of requests. As a result, the shared prefix is much cheaper than request-specific contexts.
This can reduce the hosting cost of existing LLM applications.
It may also encourage LLM customization with longer system prompts or prefix-tuning~\cite{li2021prefix} in practice.

\bibliography{custom}

\clearpage

\appendix

\section{Implementation of \relayattn}
\label{sec:impl_relay}

\textbf{Reformulation of relay fusion.} As mentioned in \cref{sec:method_relayattn}, we use the log-sum-exp trick to handle the numerical instability of the denominator in Softmax operation.
The combination coefficient for the system attention term in \cref{eq:convex_combination}, $\alpha_t^{sys}$,  is reformulated accordingly as:
\begin{equation}
\begin{split}
\alpha_t^{sys} =& \frac{\sigma_t^{1 \rightarrow s}}{\sigma_t^{1 \rightarrow l}} = \frac{\sigma_t^{1 \rightarrow s}}{\sigma_t^{1 \rightarrow s} + \sigma_t^{s+1 \rightarrow l}}\\
=& \frac{\text{exp}(\beta_t^{1 \rightarrow s})}{\text{exp}(\beta_t^{1 \rightarrow s}) + \text{exp}(\beta_t^{s+1 \rightarrow l})} \\
=& \frac{1}{1+\text{exp}(\beta_t^{s+1 \rightarrow l} - \beta_t^{1 \rightarrow s})},
\end{split}
\end{equation}
where 
\begin{equation}
\begin{split}
    \beta_t^{b \rightarrow e} = \text{log}(\sigma_t^{b \rightarrow e})
    = \text{log}(\sum_{j=b}^{e} \text{exp}(\mathbf{q}_t \mathbf{k}_j^T))
\end{split}
\end{equation}
is the log-sum-exp. 

\bigbreak
\noindent
\textbf{Implementation details.} \relayattn can be built up on existing efficient attention kernels with minimal adaptations.
For the system attention involving the system prompt of non-growing static length, we use off-the-shelf FlashAttention kernels~\cite{dao2023flashattention2}, which natively return the log-sum-exp required for computation of combination coefficients in \cref{eq:convex_combination}.
For the context attention that needs to handle the growing request-specific contexts, we use PagedAttention~\cite{kwon2023vllm} kernels for efficient memory management and modify these kernels to return log-sum-exp.
We implement a single fused kernel with OpenAI Triton~\cite{triton} for the relay fusion step, which involves multiple element-wise operations.

\section{More Information of The Datasets}

\begin{figure}[h]
    \centering
    \includegraphics[width=0.95\linewidth]{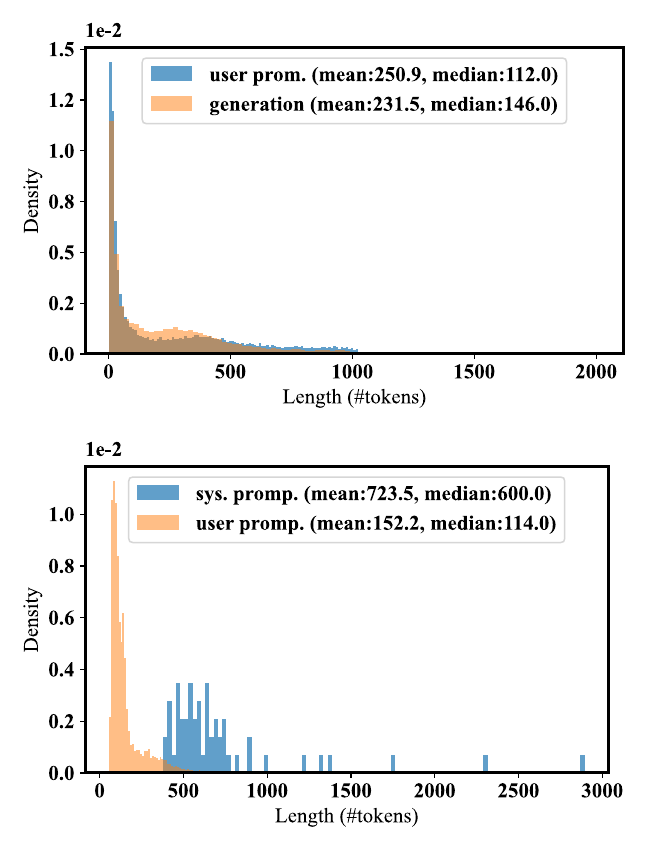}
    \caption{Distribution of the two datasets: ShareGPTv3 (top) and MMLU (bottom).}
    \label{fig:data-stat}
\end{figure}

The ShareGPTv3 dataset contains both user prompts and LLM responses.
The distributions of the length are plotted on the top of \cref{fig:data-stat}.
We use synthesized system prompts during benchmarking with this dataset.

For the MMLU dataset, we use the provided few-shot examples as system prompts and the questions as user prompts.
The generation length is set to 32 and we extract the answer in A, B, C, D as the first capital letter in the responses.
The length distributions of system prompts and user prompts are shown in \cref{fig:data-stat} bottom.

\section{Benchmark with Synthetic Workloads}
\label{sec:synthetic_workloads}

\begin{figure*}[t]
    \centering
    \includegraphics[width=1\linewidth]{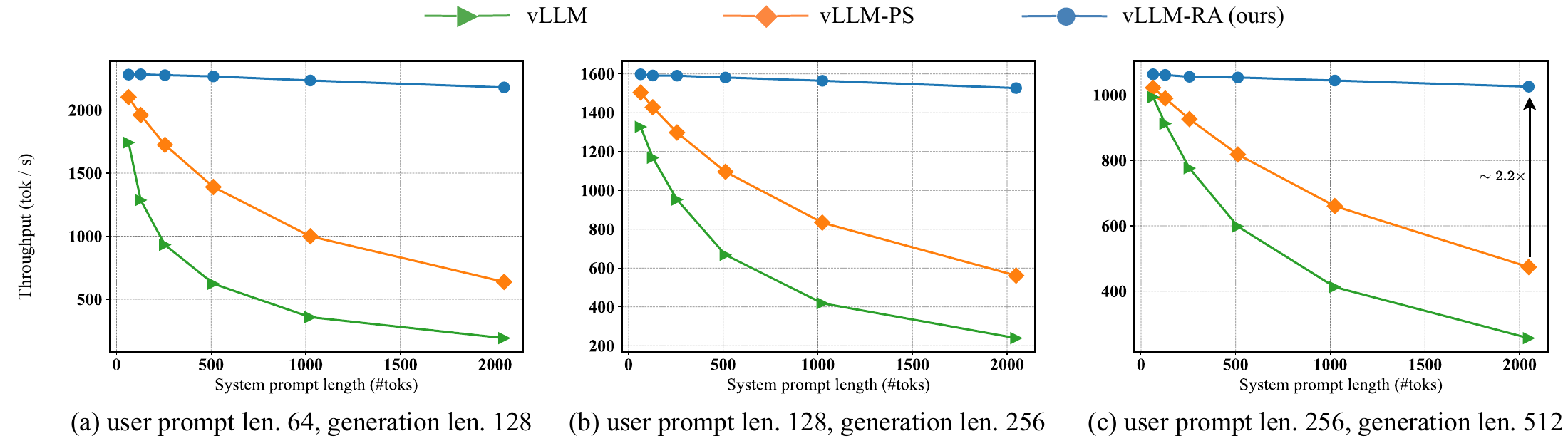}
    \caption{Throughput \wrt system prompt length with synthetic workloads.}
    \label{fig:systhetic_bench}
\end{figure*}

In the section, we benchmark the efficiency with synthetic workloads, where the user prompt length and the generation length are both fixed for all requests.
Though this is far from real-world scenarios, it is useful to test the limit of an LLM serving system because such perfectly length-aligned requests eliminate the burden of scheduling.
We adopt three combinations of user prompt length and generation length, (64, 128), (128, 256), and (256, 512) for benchmarking, and plot the trend of throughput \wrt the system prompt lenth in \cref{fig:systhetic_bench}.
Notably, in the most challenging case where the request-specific contexts have a length of $256+512=768$, \relayattn still provides an up to $2.2\times$ speedup when the system prompt length is 2048.

\section{Extension to Multi-Application Hosting}
\relayattn assumes that incoming requests share the same system prompt, which implies the serving process provides only one application. Although this is reasonable for applications (\eg, ChatGPT) with wide usage, deploying multiple applications in a single serving process is more economical in some scenarios. In such cases, it just requires some engineering efforts to support: for example, tagging each request with an application ID and implementing a scheduler to batch requests with the same application ID each time.

\end{document}